\documentclass[letterpaper, 10 pt, conference, table]{ieeeconf}  

\usepackage[noadjust]{cite}
\usepackage{diagbox}
\usepackage{amsmath,amssymb,amsfonts}
\usepackage{subcaption}
\usepackage{algpseudocode} 
\usepackage[linesnumbered,ruled]{algorithm2e}
\usepackage{xcolor,soul}
\usepackage{graphicx} 
\usepackage[font={small}]{caption}
\usepackage{booktabs}

\usepackage{epstopdf}
\newtheorem{remark}{Remark}

\IEEEoverridecommandlockouts                              

\title{\LARGE \bf
Competition-Aware Decision-Making Approach \\ for Mobile Robots in Racing Scenarios
}

\author{Kyoungtae~Ji${}^{1}$, Sangjae Bae${}^{2}$, Nan Li${}^{3}$ and Kyoungseok~Han${}^{1^*}$
\thanks{$\star$This research was supported in part by Basic Science Research Program through the National Research Foundation of Korea (NRF), funded by the Ministry of Education (NRF-2021R1A6A1A03043144); in part by the NRF grant, funded by the Korean government (MSIT) (NRF-2021R1C1C1003464)}
\thanks{$^{1}$Kyoungtae~Ji and Kyoungseok~Han are with Department of Mechanical Engineering, Kyungpook National University, Daegu 41566, Korea
        {\tt\small (\{wlrudxo644, kyoungsh\}@knu.ac.kr)}}%
 \thanks{$^{2}$Sangjae Bae is with Honda Research Institute USA, Inc. 
         {\tt\small (sbae@honda-ri.com)}}   %
\thanks{$^{3}$Nan Li is with the Department of Aerospace Engineering, Auburn University, AL, USA 
         {\tt\small (nzl0058@auburn.edu)}
}%
}

\begin{document}

\maketitle
\thispagestyle{empty}
\pagestyle{empty}

\begin{abstract}
This paper presents a game-theoretic strategy for racing, where the autonomous ego agent seeks to block a racing opponent that aims to overtake the ego agent. After a library of trajectory candidates and an associated reward matrix are constructed, the optimal trajectory in terms of maximizing the cumulative reward over the planning horizon is determined based on the level-K reasoning framework. In particular, the level of the opponent is estimated online according to its behavior over a past window and is then used to determine the trajectory for the ego agent. Taking into account that the opponent may change its level and strategy during the decision process of the ego agent, we introduce a trajectory mixing strategy that blends the level-K optimal trajectory with a fail-safe trajectory. The overall algorithm was tested and evaluated in various simulated racing scenarios, which also includes human-in-the-loop experiments. Comparative analysis against the conventional level-K framework demonstrates the superiority of our proposed approach in terms of overtake-blocking success rates.

\end{abstract}

\section{\uppercase{Introduction}}
\begin{figure}[t!]
    \centering
    \includegraphics[width=85mm]{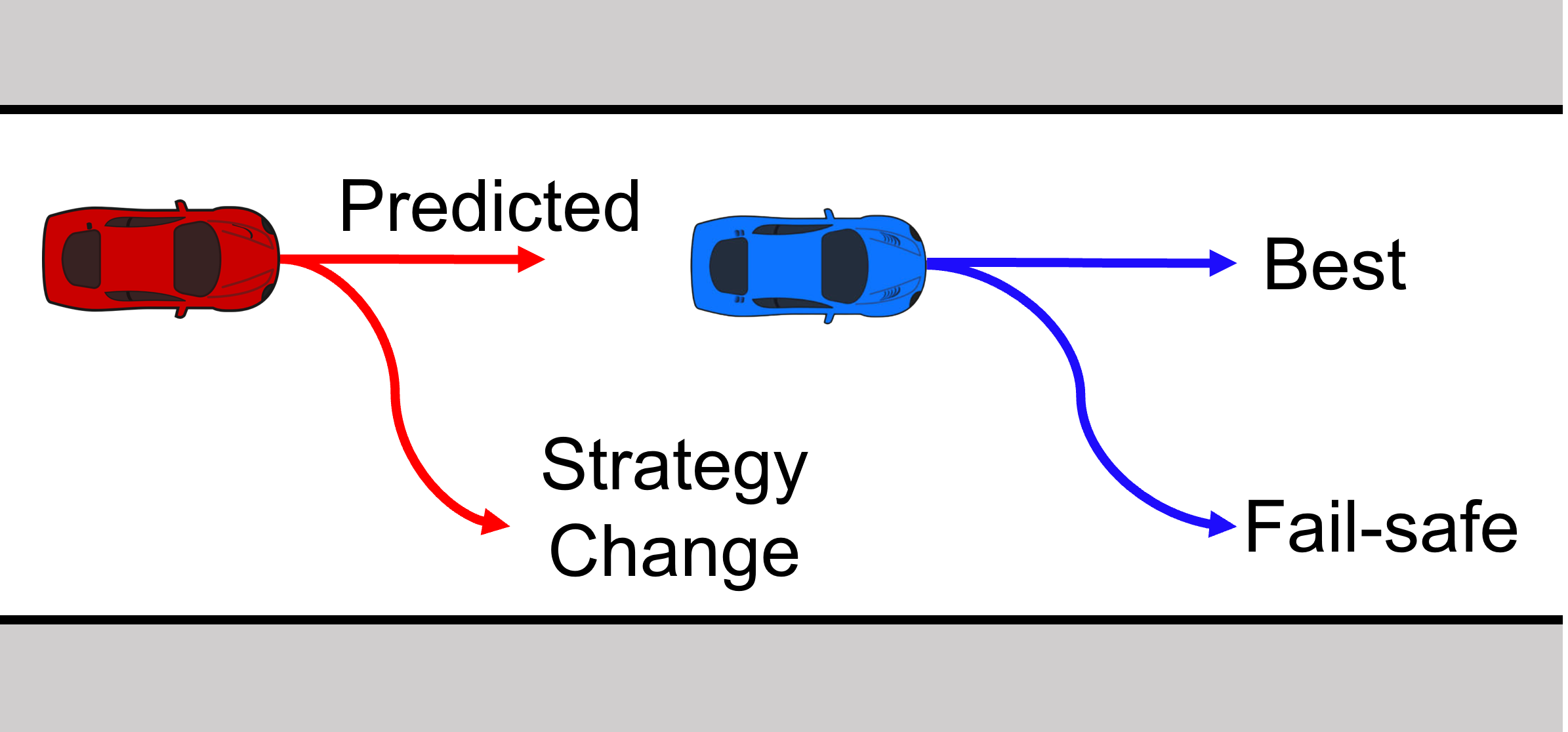}
    \caption{Example of two-players racing scenario with illustrations of possible trajectories based on the strategies.}
    \label{Scenario}
\end{figure}

In decision-making for autonomous mobile robots in environments consisting of many agents, modeling the interactions between the ego agent and surrounding agents is one of the most fundamental while challenging problems \cite{ji2020review, di2021survey, zanardi2021game}. In such a case, the reward (as a measure of performance) received by each agent depends not only on the agent's own actions but also on the actions of others nearby \cite{chandra2022gameplan}. In mobile robot racing as shown in Fig.~\ref{Scenario}, modeling interactions is even more challenging because every participating agent pursues an egocentric goal (i.e., to be the fastest and finish the course first), which can be considered a non-cooperative game \cite{bacsar1998dynamic,arbis2019game}.

Although in general mitigating the risk of collision is a paramount factor in decision-making for planning and control of mobile robots \cite{nguyen2023linear}, in competitive scenarios such as racing, overly-conservative strategies can lead to sub-optimal behavior (e.g., yield the right-of-way to an opponent) \cite{tian2022safety}. For instance, a mobile robot ahead may be able to protect its leading position by adopting a course that prevents its opponent from overtaking. However, when faced with an aggressive chaser who has been dedicated to an overtaking maneuver and coming up from behind fast, an overtaking-blocking behavior can be dangerous and result in a collision. Therefore, a prediction of the opponent's future trajectory is useful in order to react optimally \cite{laine2021multi}.

Motivated by this observation and recent studies on autonomous racing \cite{williams2017autonomous,liniger2019noncooperative,notomista2020enhancing}, in this paper, we introduce a novel game theory-inspired controller for mobile robots competitive racing scenarios. We consider a mobile robot racing where some of the robots (i.e., opponents) may be controlled by human operators and the goal of our controller for the ego robot is to beat these opponents in racing. In particular, this controller aims to protect a leading ego robot's leading position by blocking a following robot's overtaking attempt. 

Game theory has been widely employed for decision-making in interactive scenarios, where intelligent agents (called ``players'') make decisions based on predictions of each other's rational choices \cite{fisac2019hierarchical}. Game theory has better interpretability and explainability which are hardly obtained by learning methods. Various game theory-inspired strategies have been proposed for autonomous mobile robots path planning and control, such as for lane-changing \cite{zhang2019game,ji2021lane,tian2021anytime} and overtaking \cite{ji2023hierarchical}. However, a racing scenario demands a distinct approach than those for urban driving scenarios. In a racing scenario, although the robots have the same objective, their driving strategies can be different. In particular, their specific strategies depend on how they reason about and forecast each other's responses. Therefore, modeling the reasoning processes of the robots becomes an important component for developing an effective racing controller.

The level-K framework characterizes agents' strategies according to their depth of reasoning, denoted as ``level.'' It models agents' interactions based on the assumption that a level-K player makes optimal decisions while considering the other agents as level-(K$-$1) players \cite{tian2020beating,yuan2021deep,sankar2020adaptive}. In this paper, we employ the level-K framework to design a racing controller for a leading robot that is chased by a following robot who is attempting to overtake \cite{ji2023game}.

The contributions of this paper are as follows: (1) The level--K game theoretic decision-making strategy is proposed for the selection of the optimal trajectory within the competitive racing scenarios, (2) We introduce a trajectory mixing strategy to account for reasoning errors and/or strategy changes of the follower (3) To assess the effectiveness of our controller, we conducted human-in-the-loop (HIL) experiments involving several human participants. The results show that our proposed game theory-inspired controller indeed beats human-operated robots in most experiment trials. The general idea of our proposed controller may be applied to other human-robot-interaction (especially, human-robot competition) scenarios.

\section{\uppercase{Problem Definition}}

Figure~\ref{Scenario} illustrates the focused racing scenario where two players compete to outrace. 
The leading robot (in blue) is the ``ego robot'' in control. The following robot (in red) is the ``opponent'' attempting to overtake. The goal of the proposed controller is to maintain the ego robot's leading position by preventing the opponent from overtaking it. The ego robot's best trajectory is to keep straight if the opponent remains going straight. However, the opponent may detour to overtake, terminating the straight trajectory. Thus, the ego robot needs an adaptive strategy (e.g., Fail-safe in Fig.~\ref{Scenario}) to robustly outrace the opponent. To make it challenging, we allow a higher speed limit for the opponent than that for the ego robot -- thus, overtakes will be attempted.

\subsection{Differential Drive Mobile Robot Model}

We adopt the differential drive mobile robot model \cite{malu2014kinematics} of two wheels (each equipped with a motor) for the underlying controller. The kinematics reads \cite{sani2022real}:

\small
\begin{equation} \label{kinematic}
\dot{X}=\begin{bmatrix}
\dot{x}\\\dot{y}\\ \dot{\theta}
\end{bmatrix}
 = 
 \begin{bmatrix}
v \cos{\theta}\\v\sin{\theta}\\ \omega
\end{bmatrix}
=
 \begin{bmatrix}
\cos{\theta}&0\\\sin{\theta}&0\\0&1
\end{bmatrix}
 \begin{bmatrix}
v\\ \omega
\end{bmatrix},
\end{equation}
\normalsize
where the state vector $X$ represents [$x$, $y$, $\theta$] that corresponds to [longitudinal position, lateral position, orientation]; the control inputs are linear and rotational velocities $v$ and $w$, respectively.


\subsection{Polynomial Trajectory Planning}
Motivated by \cite{barghi2016trajectory,ammoun2007analysis}, the trajectories for robots are generated as a fifth-order polynomial:
\begin{subequations} \label{trajectory}
\begin{align} 
    & x(t) = a_5 t^5 + a_5 t^4 + a_3 t^3 + a_2 t^2 + a_1 t +a_0, \\
    & y(t) = b_5 t^5 + b_5 t^4 + b_3 t^3 + b_2 t^2 + b_1 t + b_0,
\end{align}  
\end{subequations}
for $t \in [0, t_T]$, where $x$ is the longitudinal position, $y$ is the lateral position, $\{(a_i, b_i), i=0,...,5\}$ are the coefficients, and $t_T$ is the planning horizon. 

The polynomial optimization determines the coefficients based on the initial and terminal states. The initial state (subscribed with $I$) is straightforwardly set to the current state:
\begin{equation}
    X_I = (x_{I}, \dot x_{I},  \ddot x_{I}, y_{I}, \dot y_{I},  \ddot y_{I}),
\end{equation}
where $\dot{\bullet}$ operates the first derivative (i.e., speed) and $\ddot{\bullet}$ operates the second derivative (i.e., acceleration). Denoted by $(x_{T}, \dot x_{T},  \ddot x_{T}, y_{T}, \dot y_{T},  \ddot y_{T})$, the terminal state relies on a target point. For a longitudinal target $x_T$, we consider the longitudinal distance with a constant acceleration $a_\text{set}$ over a fixed time period $t_T$. A lateral target $y_T$ is then set to a lateral coordinate $y_{T}$ given the lateral range of the track $y_\text{track}$. The terminal longitudinal speed $\dot{x}_T$ is integrated with the acceleration $a_\text{set}$ and the others are set to $0$, i.e., $\dot{y}_T \equiv\ddot x_{T} \equiv \ddot y_{T} \equiv 0$. Note that the terminal longitudinal speed $\dot{x}_T$ is upper-bounded by the maximum speed $v_\text{max}$. Also recall, we set the maximum speed of the opponent higher than that of the ego robot, i.e., $v_\text{max}^o > v_\text{max}^e$. Refer to Table~\ref{table:params} for the parameter settings throughout the study.

\begin{table}[]
\centering
\begin{tabular}{@{}lll@{}}
\toprule
\rowcolor[HTML]{EFEFEF} 
\multicolumn{1}{c}{\cellcolor[HTML]{EFEFEF}\textbf{Parameter}} & \multicolumn{1}{c}{\cellcolor[HTML]{EFEFEF}\textbf{Value}} & \multicolumn{1}{c}{\cellcolor[HTML]{EFEFEF}\textbf{Description}} \\ \midrule
$a_\text{set}$                                                & $\in \{-0.05, 0, 0.05\}$                                    & Acceleration, $[\frac{m}{s^2}]$                                  \\
$y_T$                                                         & $\in \{1, 1.5, 2\}$                                         & Target lateral coordinate, $[m]$                                 \\
$y_\text{track}$                                              & $[0.65, 2.35]$                                              & Track lateral range, $[m]$                                       \\
$t_T$                                                         & 5                                                           & Fixed time period, $[s]$                                         \\
$v_\text{max}^o$                                & 0.61                                                        & Speed limit of the opponent, $[\frac{m}{s}]$                                      \\
$v_\text{max}^e$                                     & 0.6                                                         & Speed limit of the ego robot, $[\frac{m}{s}]$                                     \\ \bottomrule
\end{tabular}
\caption{Parameter settings}\label{table:params}
\vspace{-5mm}
\end{table}

Note that the total number of target states corresponds to the number of possible pairs of (acceleration, target lateral coordinate), i.e., $(a_\text{set}, y_T)$ -- so does the total number of trajectory candidates. For simplicity, throughout this study, we consider three possible values of acceleration and target lateral coordinate, yielding a total of nine cases. Onward, we denote a trajectory of robot $i$ by:
\begin{equation}
    \gamma^i{(t)} = [x^i (t),y^i (t)]   \in  \Gamma^i, \quad t \in [0,t_T],
\end{equation}
where $\Gamma^i$ is the set of trajectory candidates for robot $i \in \{e, o\}$; $e$ for ego, $o$ for opponent. 

At each decision cycle, the trajectory is selected (for both the ego robot and opponent) from its candidate set $\Gamma^i$ such that a reward function is maximized. The reward function design is detailed in Section~\ref{sec:reward_function}. The selected trajectory remains the same until the current decision cycle ends.

\section{\uppercase{Interaction Modeling using Level-K framework}}

\subsection{Reward Function}\label{sec:reward_function}
In robot racing, the opponent's objective is to outrace the front robot by overtaking it. To achieve this, at time $k$, we consider three rewards:


\begin{subequations} \label{reward}
    \begin{align}
        &R_{\text{pos}}^o(k) = \sum_{j=k}^{k+N_t-1} \left (x^o(j) - x^o(k)\right ), \label{eqn:reward_lon}\\ 
        &R_{\text{rel}}^o(k) =  \sum_{j=k}^{k+N_t-1} (x^o(j) - x^e(j)),  \label{eqn:reward_rel}\\ 
        &R_{\text{block}}^o(k) = \sum_{j=k}^{k+N_t-1}\min(|y^o(j) - y^e(j)|, t_w), \label{eqn:reward_block}
        \end{align}
\end{subequations}
where $N_t$ is the prediction horizon of the trajectory (i.e., $t_{T}/$sample time) and $t_w$ is the robot track width (set to  $0.3 [m]$). 

Each reward is described as follows:
\begin{itemize}
    \item In Eqn.~\eqref{eqn:reward_lon}, the position reward $R_\text{pos}^o(k)$ formulates the progress in the longitudinal direction with respect to the initial position. This encourages a higher speed.
    \item In Eqn.~\eqref{eqn:reward_rel}, the relative distance reward $R_\text{rel}^o(k)$ formulates the opponent's longitudinal position with respect to the ego robot's. This encourages the overtakes.
    \item In Eqn.~\eqref{eqn:reward_block}, the block reward $R_\text{block}^o(k)$ formulates the inter-vehicle gap in the lateral direction. This encourages lateral deviations, while the reward is upper-bounded by $t_w$ (to ensure minimum effort to overtake).
\end{itemize}


The opponent's holistic reward is a convex combination of each reward in Eqn.~\eqref{reward}:
\small
\begin{equation}
    \mathcal{R}\mathit{}^o(\gamma^e(k),\gamma^o(k)) = w_1 R_{\text{pos}}^o(k) + w_2 R_{\text{rel}}^o(k) + w_3 R_{\text{block}}^o(k),
\end{equation}
\normalsize
where $w_{1,2,3} = [1,0.5,1]$ represent positive weighting coefficients that reflect the relative importance of each reward component. We tuned the weights through the Human-in-the-loop experiments to imitate naturalistic behaviors.

The competitive two-player game can be formulated as a zero-sum game \cite{vamvoudakis2023game}. We thus formulate the ego robot's reward as the negative reward (i.e., cost) of the opponent:
\begin{equation}
    \mathcal{R}^e(\gamma^e(k),\gamma^o(k)) = - \mathcal{R}^o(\gamma^e(k),\gamma^o(k)).
\end{equation}

With the reward $R_\text{block}$, the opponent may choose a trajectory that avoids collisions. In contrast, the ego robot may choose a risky trajectory that brings it closer to the opponent\footnote{This might result in collisions, which we further discuss in Section~\ref{sec:conclusions}}. This competitive dynamics is aligned with the principle of the zero-sum game \cite{maschler2020game}.


\subsection{Decision-Making Strategy based on Level-K Framework}
In interactive scenarios, it is common for agents to make decisions while considering the predicted actions of others. In this study, we model the strategies of each agent based on the level-K framework to identify the depths of reasoning. 

First, the level-0 player is assumed to be the naive player who reacts solely to the available information without accounting for interactions. We specify the level-0 strategy for each agent as follows:
\begin{subequations} \label{10}
\begin{align}
        \gamma^{e,*}_0 \in \underset{\gamma^e \in \Gamma^e}{\text{argmax}} \,\,\mathcal{R}\mathit{}^e(\gamma^e),\\
        \gamma^{o,*}_0 \in \underset{\gamma^o \in \Gamma^o}{\text{argmax}} \,\,\mathcal{R}\mathit{}^o(\gamma^o),
\end{align}
\end{subequations}
where $\gamma^{i} \in \Gamma^{i}, i=\{e, o\}$ represent the trajectory candidates for the ego robot and opponent, and subscript 0 denotes the level-0. Here, we omit the time instance $k$ for brevity.

Based on this, the level-0 agent chooses a trajectory to maximize the cumulative reward while assuming that the other agents as the stationary obstacle. 

After defining the level-0 strategy, the level-K strategy can be defined while assuming the other agents are modeled by level-(k$-$1) reasoning:
\begin{subequations} \label{11}
    \begin{align}
            \gamma^{e,*}_k \in \underset{\gamma^e \in \Gamma^e}{\text{argmax}} \,\,\mathcal{R}\mathit{}^e(\gamma^{o,*}_{k-1},\gamma^e),\\
        \gamma^{o,*}_k \in \underset{\gamma^o \in \Gamma^o}{\text{argmax}} \,\,\mathcal{R}\mathit{}^o(\gamma^o,\gamma^{e,*}_{k-1}),
    \end{align}
\end{subequations}

This strategy involves a reward matrix (a.k.a. payoff matrix \cite{maschler2020game}), where the rewards are computed based on the above trajectory combinations. To solve the \eqref{10} and \eqref{11}, we conduct the reward matrix for level-0 as $\mathbb{R}^{N\times1}$ and the level-K as $\mathbb{R}^{N \times N}$ for $N$ trajectory candidates. The best trajectory for the level-K strategy can be obtained using the ``max()" function to a specific row or column of the reward matrix.

Existing literature suggests that the common depth of reasoning in interactive situations falls within the range of 0 to 2 \cite{li2017game}. Consequently, we consider player levels ranging from 0 to 2. Specifically, the best trajectory for the opponent is computed at levels 0 to 2, while the best trajectory for the ego robot is determined at levels 1 to 3.

\section{\uppercase{Overtake-blocking Controller}}
\begin{figure}
    \centering
    \includegraphics[width=1\linewidth]{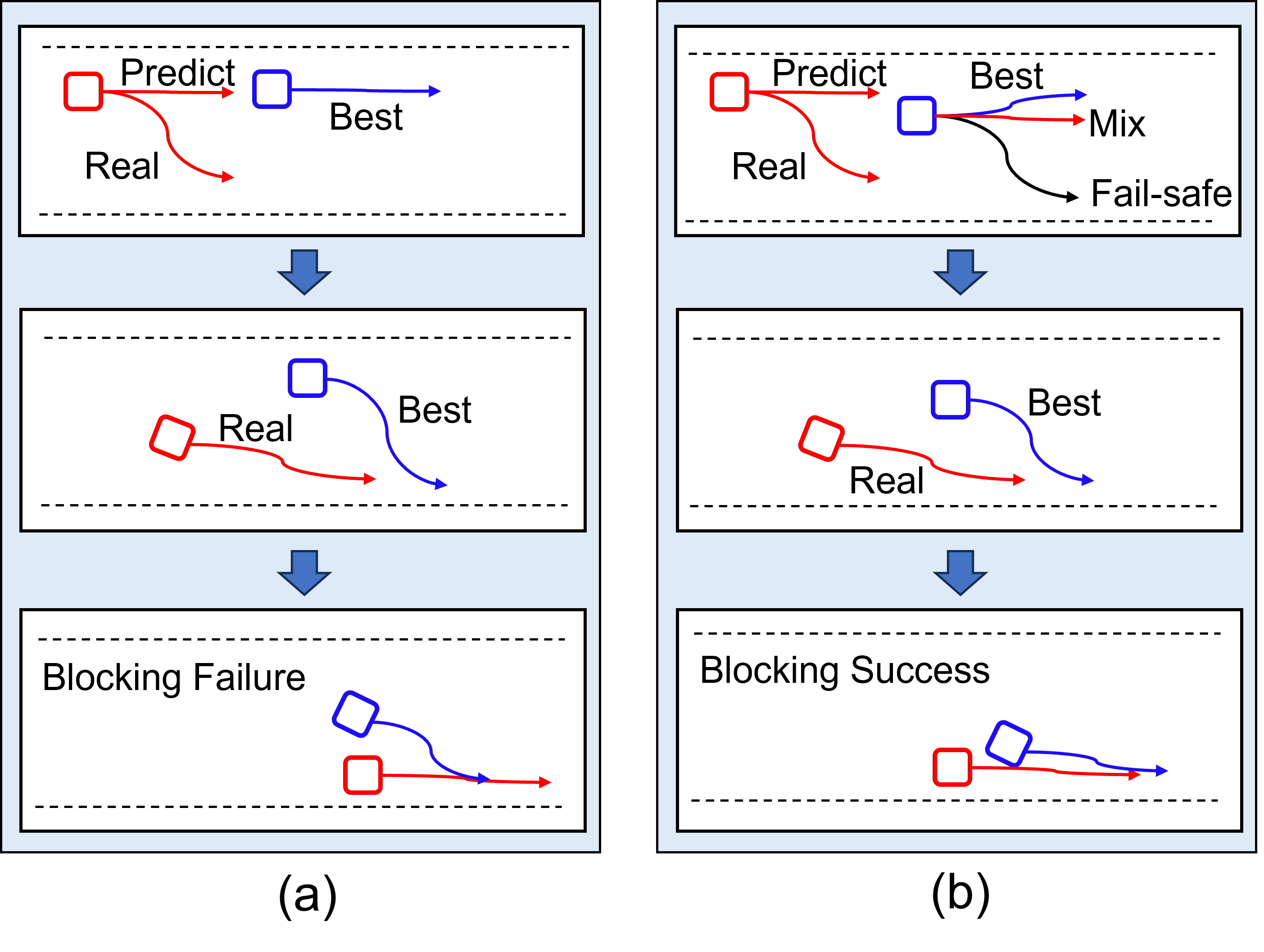}
    \caption{The example of the opponent's sudden level change. (a) The ego robot is overtaken by the opponent due to the reaction delay. (b) The ego robot blocks the overtaking attempt by following a mixed trajectory.}
    \label{Balance}
\end{figure}

Within the level-K framework, the estimation of other players' levels is crucial for selecting appropriate actions. In typical urban driving scenarios \cite{liu2022interaction}, it can be assumed the constant level of interactive agents during the short-term interaction period. However, in competitive scenarios, rational players have the ability to adapt their strategies in order to outperform their opponents. In other words, the agents frequently change their level depending on the behaviors of other agents \cite{jung2021game}. For instance, when the robot behind suddenly changes the level and attempts to overtake, the conventional level-K framework can fail to block due to the estimation delay and physical limitation of the robot as shown in Fig.~\ref{Balance}(a) \cite{onieva2010overtaking}. As a solution to this, we propose the adaptive trajectory mixing approach. As shown in Fig.~\ref{Balance}(b), the ego robot mixes the best trajectory with the fail-safe trajectory which is the best response for the opponent's least likely trajectory. That is, instead of following the best trajectory, the ego-robot follows the mixed one to take into account the sudden level change of the robot behind. This allows the ego robot to be wary of the opponent's sudden level change and can block the overtaking attempt under more relaxed conditions compared to the case in Fig.~\ref{Balance}(a).


\subsection{Opponent's Level Estimation}
The level of the opponent is updated as follows \cite{li2018game}:
\begin{align} \label{level}
    k^* &\in \underset{k \in \{0,1,2\}}{\text{argmin}} | \gamma^o_{real}(-N_p: 0) - \gamma^{o,*}_{k,old}(-N_p: 0)|, \nonumber\\
P^{o}(k^*) &\leftarrow P^{o}(k^*) + \Delta P, \\
\quad \, \, \,  P^{o}(k) &\leftarrow P^{o}(k) \Big/  \sum\nolimits_{k =0}^2(P^{o}(k)), \: k \in \{0,1,2\} \nonumber
\end{align}
where $\gamma^{o}_{real}$ is the real past trajectory of the opponent from $N_p =5$ before to current time step $0$, $\gamma^{o,*}_{k,old}$ is the predicted trajectories for the level-K opponent from the previous decision cycle, and $P^o(k)$ denotes the belief that opponent follows the level-$k$ strategy. 

To update the belief, we identify the opponent level whose trajectory most closely matches the real trajectory over the specified time horizon. We then increment the belief of that level by a tunable update value $\Delta P = 0.5$. Subsequently, we normalize all belief values to ensure their sum equals 1.

The ego robot utilizes these beliefs to infer the opponent's level. Specifically, the ego robot assumes that the opponent's level corresponds to the level with the highest belief value and selects a trajectory one level higher than the opponent's one. Initially, even beliefs are assumed for each level, and the above estimation process is conducted after $N_p$ step. In cases where multiple belief values are equal, level inference priority is assigned to the lowest level.

\subsection{Adaptive Trajectory Mixing Approach}

To create this adaptive trajectory, we introduce the concept of level change potential $P_c^o \in [0, P_c^{\mathrm{lim}}]$ which means the degree of wary about the opponent's level change. If $P_c^o=0$, the opponent's level change is not considered. In contrast, if $P_c^o$ takes the value, it indicates the confidence degree of the opponent's level change. The level change potential is updated at each decision cycle as follows:
\begin{subequations}\label{change}
\begin{align}  
P_c^{o} &\leftarrow P_c^{o} + \Delta P_c^o, \\
P_c^{o} &\leftarrow \max(\min(P_c^o,P_c^{\mathrm{lim}}),0)
\end{align}
\end{subequations}
where $\Delta P_c^o$ takes a negative value when the level change is observed while taking a positive value if the opponent maintains the level, and $P_c^{\mathrm{lim}} = 0.2$.  

Based on this, the level change potential approaches 0 if the opponent changes the level, and it increases if the opponent holds the current level.

\begin{algorithm}[t!] 
	\caption{Adaptive Trajectory Mixing Process} \label{alg1}
	Input $ \gamma^{e,*}_{N^e}, P^o{(N^o)}, P^{o}_{old}{(N^o)},P_c^o$

    \eIf{$ \underset{k \in N^o}{\mathrm{argmax}}\: P^o(k) \neq \underset{k \in N^o}{\mathrm{argmax}}\: P^o_{old}(k)$}{
    $\Delta P_c^o = -P_c^{lim}$
    }{
$\Delta P_c^o = 0.05$
    }
Update $P_c^o$ through \eqref{change}

$k^* \leftarrow \underset{k \in N^o}{\text{argmax}} \:P^o(k) $

$k^{\text{fail}} \leftarrow \underset{k \in N^o}{\text{argmin}} \:  P^o(k) $

$\gamma^{e,*} \leftarrow (1- P^o_c) \gamma^{e,*}_{(k^*+1)} + P^o_c \gamma^{e,*}_{(k^{\text{fail}}+1)}$

Output $\gamma^{e,*}$
\end{algorithm}

For better understanding, we present the detailed adaptive trajectory mixing process in Alg.~\ref{alg1}. We represent the set of the levels for the ego robot and opponent as $N^e = \{1,2,3\}, N^o = \{0,1,2\}$. In line 1, the inputs are the ego robot's best trajectories depending on level $N^e$, the belief of the opponent's level for current and previous decision cycle $ P^o{(N^o)}, P^{o}_{old}{(N^o)}$, and $P_c^o$. First, the ego robot checks the level change of the opponent by comparing the estimated opponent's level using the current and previous beliefs ($P^o(k), P^0_{old}(k)$).

If the opponent changes the level, then give the update value $\Delta P_c$ to the negative value same as the upper limit, and 0.05 otherwise (lines 2-6). Therefore, if the estimated opponent's level changes, $P_c^o$ becomes zero, otherwise, $P_c^o$ increases incrementally over time (line 7). Finally, the ego robot infers the level of the opponent (line 8) and prepares for the opponent's future trajectory by considering the lowest belief level trajectory of the opponent (line 9). Consequently, the ego robot's final trajectory is a combination of the best trajectories for the current state and the fail-safe trajectory to consider the opponent's least likely level, based on $P_c^o$ (line 10). 

This update mechanism allows the ego robot to follow the opponent's updated level trajectory when a level change occurs, while also preparing for a level change if the opponent maintains the current level. The upper limit of the level change potential $P_c^{\mathrm{lim}}$ is defined to prevent the ego robot from neglecting to block the current level trajectory due to excessive vigilance.

\remark{Preparing for level changes in other agents is not feasible in the game with the action space consisting of discrete actions. It is because the blending may not be natural, for example, blending the steering angle or pedal angle. However, with trajectory candidates, the x-y positions at the same time steps can be blended using the proportion determined by $P_c^{o}$.}

\textcolor{red}{}

\subsection{Trajectory Tracking Controller}
The model predictive control (MPC) is designed to follow the selected trajectory \cite{worthmann2015model, han2020model}: 

\small
\begin{align}
\min_{u(0)...u(N_p-1)} J =  &\sum_{k=0}^{N_p-1} ||{X}(k+1) - \gamma^{ref}||^2_Q + ||{u}(k) - {u}^{ref}||^2_R\\
\textrm{s.t.}\,\, &\eqref{kinematic}, {X}(0) = {x}_0 \nonumber\\
  &{u}(k) \in \mathcal{U}, \: \forall k \in [0:N_p-1]\nonumber\\
  &{X}(k) \in \mathcal{X}, \:  \forall k \in [0:N_p-1]. \nonumber
\end{align}
\normalsize
where $N_p$ is the prediction horizon, $X$ and $u$ are augmented states in \eqref{kinematic} and the control inputs $v, w$, $\gamma^{ref}$ and $u^{ref}$ are the references for the augmented state and control inputs. $Q$ and $R$ are the weighting metrics, and $\mathcal{U}$ and $\mathcal{X}$ are control and state constraints.


\begin{figure}
    \centering
    \includegraphics[width=1\linewidth]{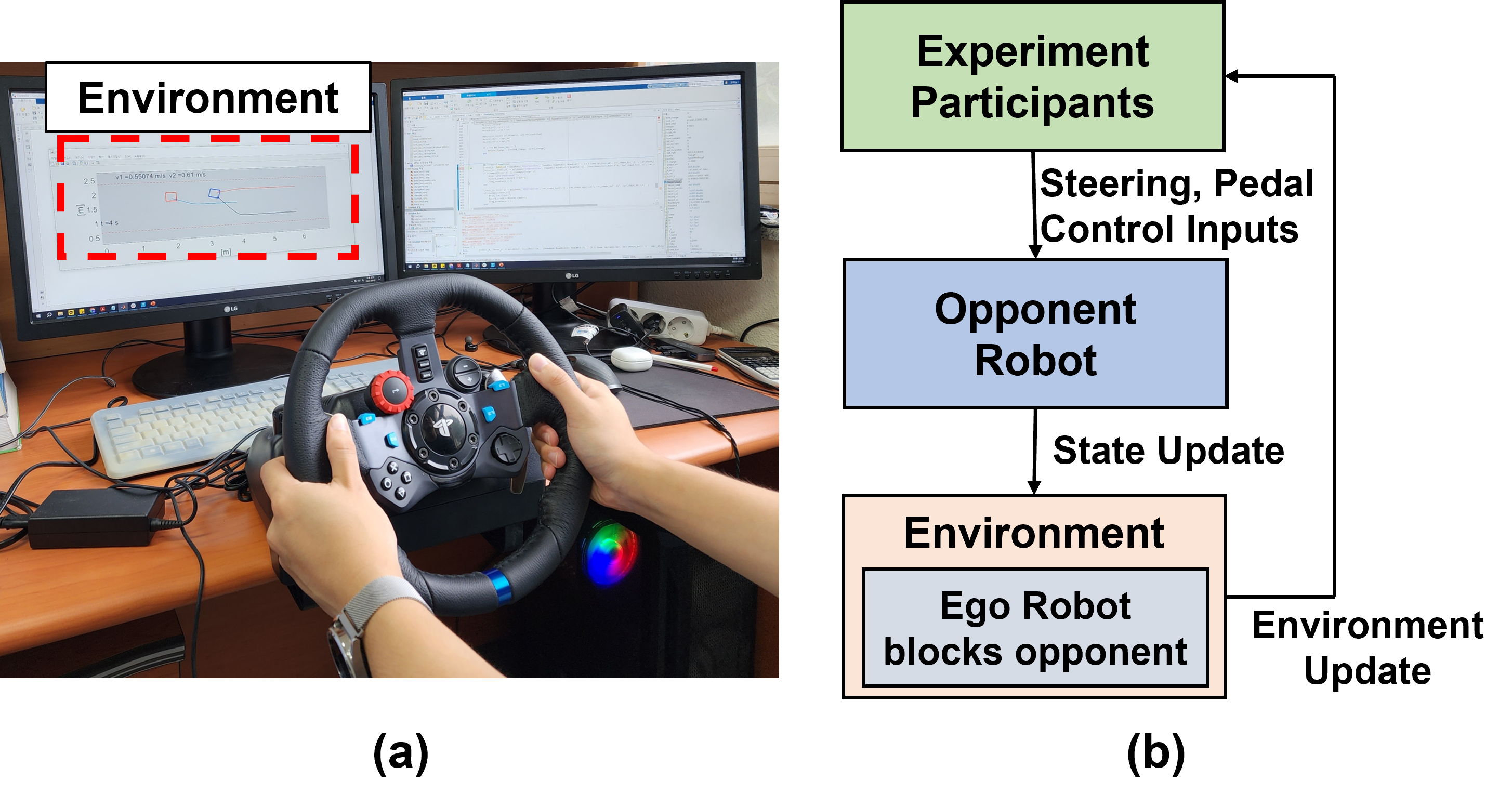}
    \caption{ Experimental setup. (a) The experiment involving human participants to control the opponent robot, (b) The process of the experiment.}
    \label{setup}
\end{figure}

\section{\uppercase{Experimental Results}}
\subsection{Environment Setup}
In this section, we present the experimental results of our study. The simulation settings included a sample time of 0.2 seconds and a decision cycle of 1 second, and the decision-making of the proposed controller took an average of 3.82 ms. The test platform is built on Matlab R2021a under the PC with Intel i5-9500 CPU, and RAM 16 GB.

In competitive game scenarios, accurate modeling of the opponent is pivotal for ensuring reliable verification. To this end, we have designed multiple opponent models tailored to various scenarios as follows:

\begin{enumerate}
    \item Following the preassigned constant-level strategy 
    \item Following the random trajectory for each sample time
    \item Controlled by human operators
\end{enumerate}

\begin{figure} [t!]
    \centering
    \includegraphics[width=0.95\linewidth]{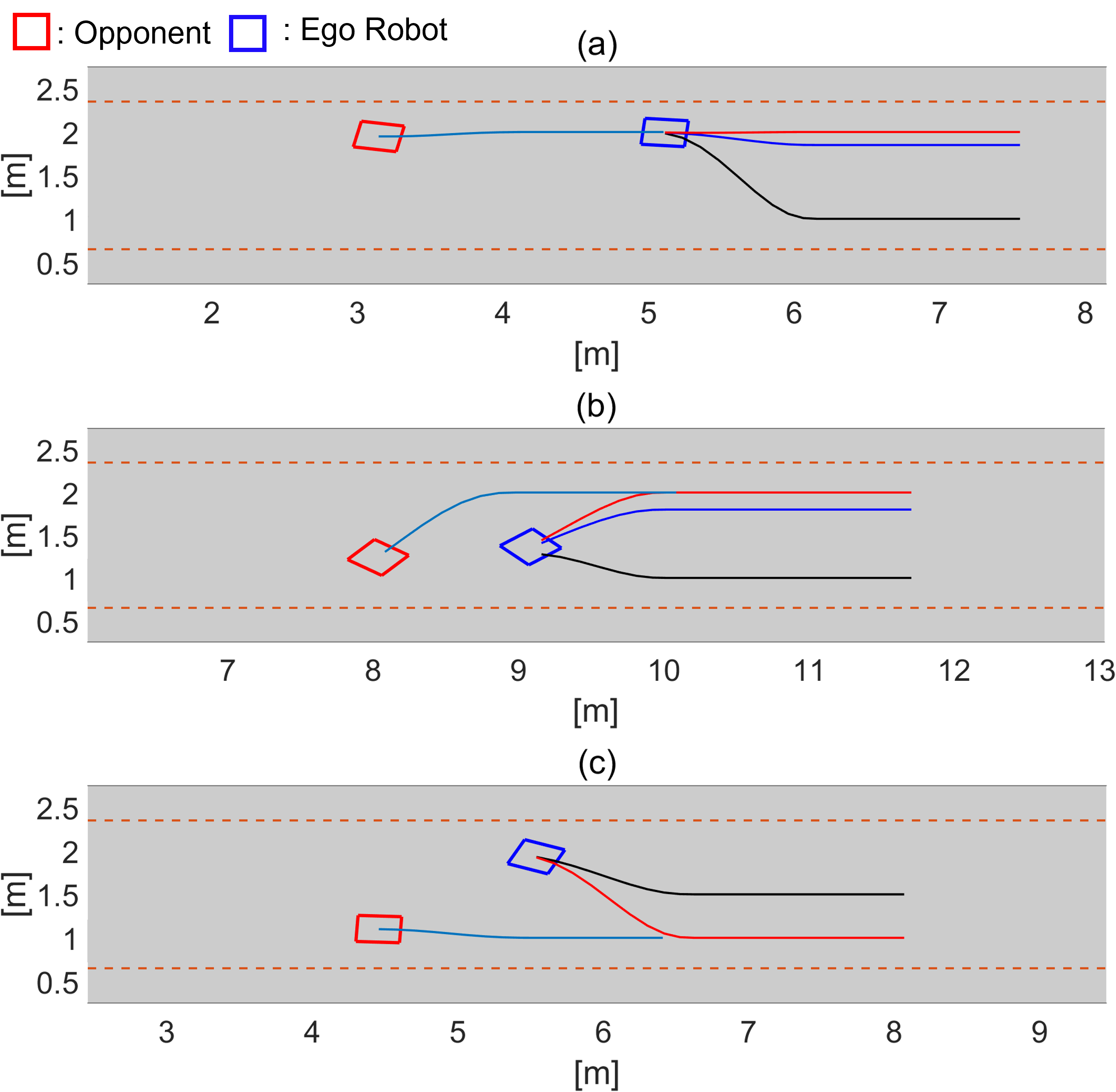}
    \caption{The representative interacting case of the two robots racing scenario, where the blue square is the ego robot controlled by our approach, and the red square is the opponent robot controlled by three different approaches. The lines in front of the robots are planned trajectories (red: best, black: fail-safe, blue: mixed). (a) The ego robot blocks in the way of the opponent. (b) The ego robot follows the mixed trajectory to block the opponent's predicted trajectory. (c) The ego robot follows the trajectory to block the opponent who is already far from the ego robot, the mixed trajectory overlaps with the best trajectory.}
    \label{Case}
\end{figure}

First, we designed the opponents following randomly assigned constant levels during the simulation. In the second case, the opponents follow the random trajectory candidate at each sample time without a predetermined strategy. 
Lastly, we built HIL experiments using a Logitech G29 driving wheel and pedals, as depicted in Fig.~\ref{setup}.

The tests were conducted 200 times for each scenario, with the random initial position assigned to the opponent up to 2 meters from the robot ahead. Additionally, the scenario involving human operators was performed 20 times for each of the 10 operators to enhance reliability.

We extended our validation with comparative analysis against the conventional level-K framework, which always follows the best trajectory according to the estimated opponent's level without considering the opponent's level change. In each simulation, we categorized the outcomes as ``overtaking success'' when the opponent's longitudinal position surpasses that of the ego robot without a collision within 60 seconds, and ``blocking success'' otherwise. The simulation results are presented in terms of the number of blocking the overtaking of the opponent.

\begin{figure}[t!]
	\centering
	\begin{subfigure}{.5\textwidth}
	\centering
    \includegraphics[width=1.0\linewidth]{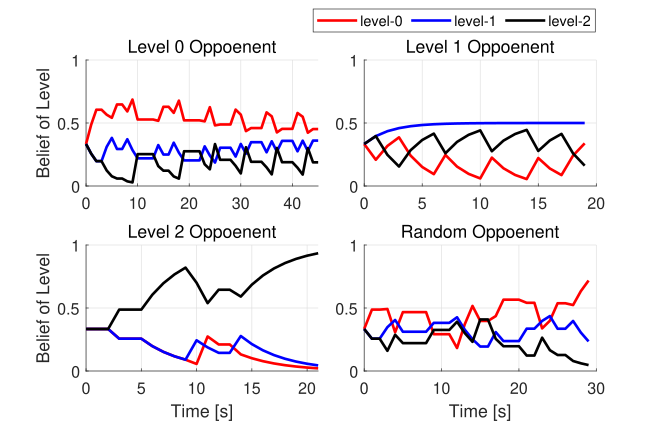}
    \caption{The estimated belief of the level of the opponent}
	\end{subfigure}
	\begin{subfigure}{.4\textwidth}
    \includegraphics[width=1.0\linewidth]{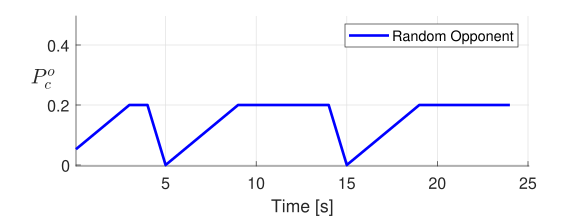}
    \caption{The potential for level change of the opponent}
    \label{Change}
	\end{subfigure}
 	\caption{The estimated belief of the opponent's level for representative scenarios.}
  \label{TotalLevel}
\end{figure}

\subsection{Test Results}
First, we analyze the representative cases of the interaction between the proposed controller and level-K opponent as shown in Fig.~\ref{Case}. The blue square is the ego robot, and the red square is the opponent. The lines in front of the robots are planned trajectories. The red line is the best trajectory for the estimated level of the opponent, and the black line is the fail-safe trajectory for the opponent's lowest belief trajectory. The blue line is the mixed trajectory that mixed the fail-safe trajectory to prepare for the opponent's sudden level change.

Fig.~\ref{Case}(a) shows that the ego robot blocks in the way of the opponent and keeps the trajectory in the belief that the opponent will maintain in current lateral position. This is because the ego robot believes that the opponent assumes that the ego robot will change lanes (same as level-1). Fig.~\ref{Case}(b) shows that the ego robot targets the left side of the track to block the opponent's future trajectory. In this case, the ego robot believes the opponent will go to the left side of the track even if the current position is near the right side of the track. In both cases, the ego robot keeps the belief of the opponent's level, so follows the adaptive trajectory (blue lines) to prepare for the sudden level change of the opponent. In contrast, Fig.~\ref{Case}(c) shows the case that the opponent is already far from the ego robot in terms of the lateral distance, and the ego robot follows the trajectory to block the progress of the opponent. In Fig.~\ref{Case}(c), the adaptive trajectory overlaps with the best trajectory of the estimated level of the opponent. This is because of the lower value of level change potential to avoid blocking failure for the opponent's overtaking when following the opponent's changed level.

\begin{figure}
    \centering
    \includegraphics[width=75mm]{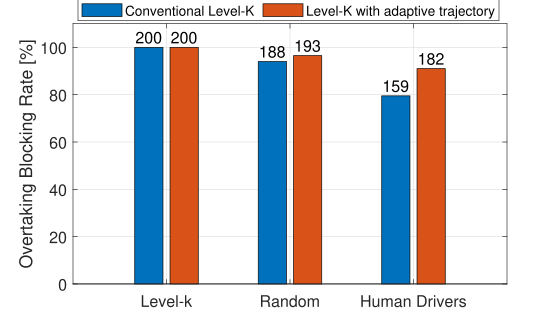}
    \caption{The test results in terms of overtaking blocking rates. Every scenario was performed 200 times with random initial conditions. In the human operator scenarios, 10 participants are involved in 20 experiments independently. }
    \label{Result}
\end{figure}

The examples of the level estimation for the different opponent models are shown in Fig.~\ref{TotalLevel}(a). The estimated beliefs of the level for level-0,1,2 opponents are straightforward. However, the estimated level of the random opponent fluctuates since the opponent has a strategy that level-K did not expect (choose random trajectory without rationality). In this case, as shown in Fig.~\ref{TotalLevel}(b), the potential for level change maintains to upper limit $P_c^{\text{lim}}$ when the opponent's level is estimated to same and becomes 0 at the instance of the estimated opponent's level change. The level-0,1,2 opponent's potential for level change maintains $P_c^{\text{lim}}$ and we omit it in the figure for brevity.

The overall results of the experiment are shown in Fig.~\ref{Result}. First, both approaches block the overtaking of constant level-0,1,2 opponents perfectly. In the random opponent scenario, the conventional level-K controller blocks overtake for 94\%, and our approach blocks overtaking attempts for 96.5\%. The percentage of blocking the human operators' overtaking in both controllers is 79.5\% and 91\%. In conclusion, the performance of our approach outperforms the conventional level-K controller in terms of success rates.

\section{CONCLUSIONS}\label{sec:conclusions}
This paper presents a game-theoretic competition-aware strategy to block a following opponent in a racing scenario. The proposed approach employs a level-K framework to account for the opponent's rationality levels and strategy changes. Experimental results demonstrated the efficacy of the approach in blocking overtaking maneuvers, outperforming the conventional level-K controller, particularly in scenarios involving random opponents and human operators. This approach provides a promising strategy for autonomous vehicles to compete effectively in dynamic and unpredictable racing scenarios. {Simplifications made in this work lead to the limitations and corresponding future work: (i) Safety was not considered paramount to design an effective strategy, which results in collisions when the opponent is irrationally aggressive; (ii) The number of trajectory candidates was limited to few to secure real-time efficiency, which might lead to jerky behaviors in practice. (iii) The validation environment was relatively simple (with only one opponent). The future works include addressing the aforementioned limitations. Also, integrating the approach into a physical system and experiments with a scaled robot are to be conducted.


\bibliographystyle{IEEEtran}

\bibliography{ref}

\end{document}